\documentclass[sigconf]{acmart}
\usepackage{amsmath}
\usepackage{url}
\usepackage{graphicx, caption, subcaption}
\usepackage{float}
\usepackage{blindtext}
\theoremstyle{remark} \newtheorem{mydef}{Definition}
\newcommand{\tabincell}[2]{\begin{tabular}{@{}#1@{}}#2\end{tabular}}

\begin{document}
\settopmatter{printacmref=false}
\renewcommand\footnotetextcopyrightpermission[1]{} 
\pagestyle{plain} 
\title{Long-term Joint Scheduling for Urban Traffic}



%


\author{Xianfeng Liang$^{1}$, Likang Wu$^{1}$, Joya Chen$^{1}$, Yang Liu$^{1}$, \and Runlong Yu$^{1}$,  Min Hou$^{1}$, Han Wu$^{1}$, Yuyang Ye$^{1}$, Qi Liu$^{1}$, Enhong Chen$^{1}$}
\affiliation{\institution{\textsuperscript{1}Anhui Province Key Laboratory of Big Data Analysis and Application, University of Science and Technology of China}}
\email{{zeroxf,wulk,chenjoya,liuyang0,yrunl,minho,wuhanhan,yeyuyang}@mail.ustc.edu.cn {qiliuql,cheneh}@ustc.edu.cn}

\begin{abstract}
Recently, the traffic congestion in modern cities has become a growing worry for the residents.
As presented in Baidu traffic report~\cite{baidu}, the commuting stress index $\footnote{The commuting stress index is the ratio of actual travel time to unobstructed travel time in the morning and evening rush hours of working day.}$, has reached surprising 1.973 in Beijing during rush hours,
which results in longer trip time and increased vehicular queueing.
Previous works have demonstrated that by reasonable scheduling, e.g, rebalancing bike-sharing systems \cite{ghosh2016robust} and optimized bus transportation \cite{wang2017data}, the traffic efficiency could be significantly improved with little resource consumption.
However, there are still two disadvantages that restrict their performance:
(1) they only consider single scheduling in a short time, but ignoring the layout after first reposition,
and (2) they only focus on the single transport. However, the multi-modal characteristics of urban public transportation are largely under-exploited.
In this paper, we propose an efficient and economical multi-modal traffic scheduling scheme named JLRLS based on spatio-temporal prediction,
which adopts reinforcement learning to obtain optimal long-term and joint schedule. In JLRLS, we combines multiple transportation to conduct scheduling by their own characteristics, which potentially helps the system to reach the optimal performance.
Our implementation of an example by PaddlePaddle is available at \url{https://github.com/bigdata-ustc/Long-term-Joint-Scheduling}, with an explaining video at \url{https://youtu.be/t5M2wVPhTyk}.

\end{abstract}




\maketitle

\section{INTRODUCTION}

The rapid expansion of urban traffic, with the slow growth of traffic resources, has led to the serious and growing traffic congestion.
According to a Baidu traffic report \cite{baidu}, the commuting index has raised to 1.973 during rush hours in Beijing.
Traffic congestion poses a great threat to traffic safety and also brings losses to the urban economy.
Fortunately, previous works have proved that reasonable traffic scheduling can improve the traffic efficiency with less consumption.
In ~\cite{wang2017data}, a data-driven optimization algorithm for bus system is proposed, which reduces the average waiting time of citizens.
A bike-sharing scheduling system~\cite{ghosh2016robust} proposes an online and robust framework to minimize the loss of customers.

\begin{figure}[h]
  \centering
  \includegraphics[width=\linewidth]{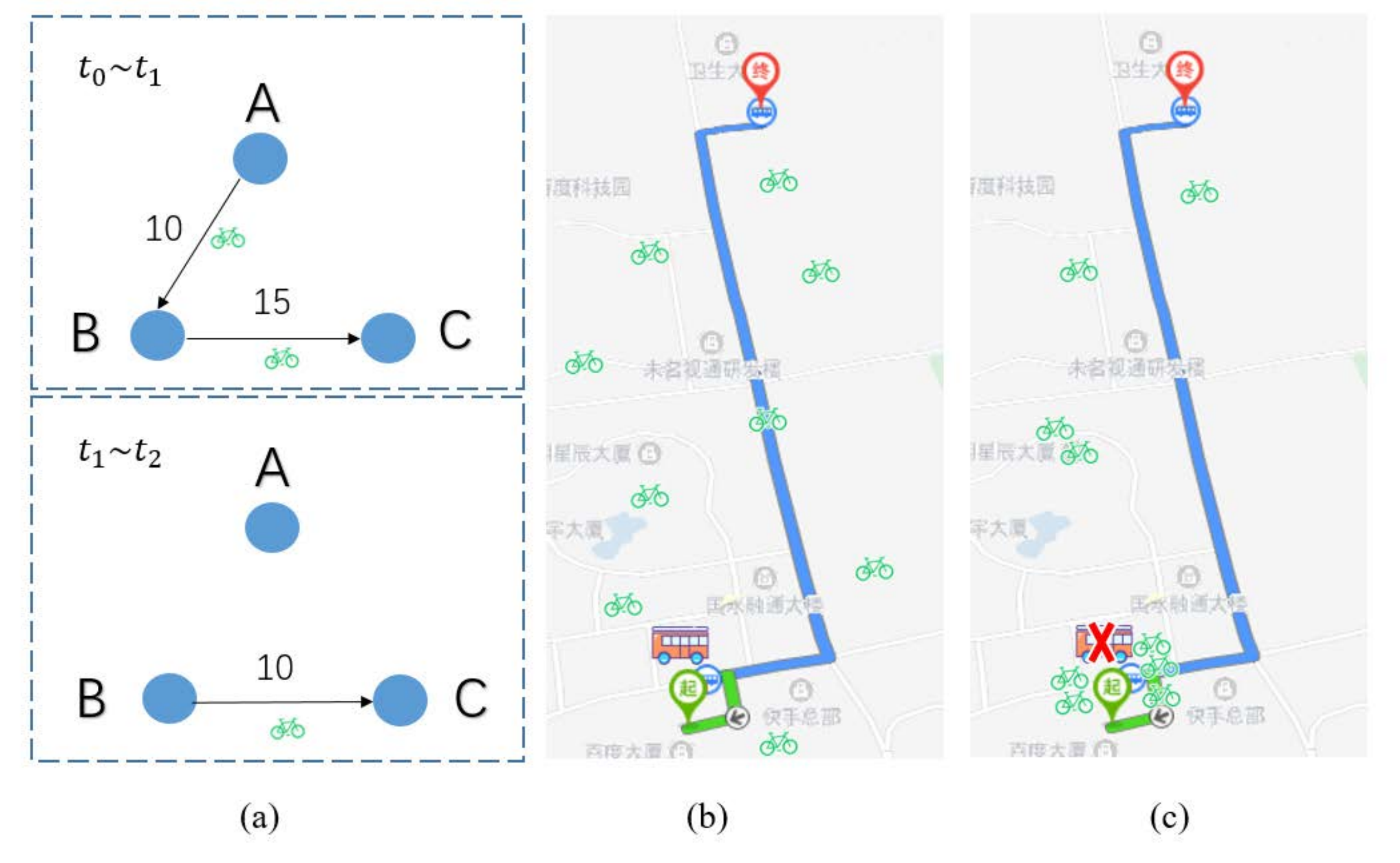}
  \caption{The Case of Traffic System}
 \label {fig:case}
\end{figure}

However, there are still two disadvantages that restrict the further improvement of efficiency:
(1) typical bike scheduling, e.g, Liu et al ~\cite{liu2016rebalancing},  proposes a hierarchical
optimization model for rebalancing by exploring multi-source data. But they only consider single planning in a short time, ignores analyzing the situation after first planning.
A common example is in Figure \ref{fig:case} (a). There are three bike-sharing stations ($A$, $B$ and $C$) without available bikes.
We assume that 10 customers ride from $A$ to $B$, and 15 customers ride from $B$ to $C$ during $t_0 \sim t_1$. From $t_1$ to $t_2$, 10 customers ride from $B$ to $C$.
And only 10 bikes can be dispatched before $t_0$.
According to the greedy strategy, during $t_0 \sim t_1$, $B$ will be assigned to 10 bikes. But with consideration of the dynamic flow, 10 bikes should be moved to A, as it can finally serve 20 customers.
Therefore, if the secondary planning can be carried out, the local greedy problem could be alleviated.
(2) Current works  ~\cite{ghosh2016robust, wang2017data, liu2016rebalancing, li2018dynamic} only adopt single modes of transport, while ignoring the multi-modal characteristics of urban public transport.
For example, as shown in Figure \ref{fig:case} , (b) indicates the normal operation of the bus, and (c) indicates that when the bus is unavailable, the system can automatically move the shared bikes to replace the bus.

Joint scheduling for long periods is usually difficult.
Human activities are social and uncertain, which may result in an extremely imbalance between supply and demand of the traffic.
Moreover, it increases the difficulties of traffic scheduling:
(1) The accuracy of the predication of future demand should be as high as possible, as it directly affects the subsequent optimization.
(2) Traditional scheduling is complex and can not be applied to large-scale problems.
Furthermore, traditional algorithms use MIP to solve optimization, which only suits for obtaining a solution in special environments.
(3) Joint scheduling depends on setting rules manually, which may produce greedy and short-sighted strategies.

In this paper, we investigate multi-modal transposition carefully, and discover that different traffic modes can be scheduled complementarily to improve efficiency.
It should be noted that the characteristics of different transports are different. For example, bike-sharing scheduling is flexible, so it could be affected by other transports easily. Therefore, bike-sharing scheduling is suitable for scheduling with others. Some traffic scheduling, such as bus scheduling is less affected by other transports, and thus relatively fixed.To consider the interaction between different traffic systems, we design a global scheduling method that can schedule both types of traffic at the same time, so that the flexible traffics can be dispatched in coordination with the fixed traffic to achieve the global optimality.
Specifically, for the most common transports — bus and bike, a joint traffic scheduling framework named JLRLS based on reinforcement learning is proposed. JLRLS incorporates the bike flow into consideration, which helps to avoid local greed. Meanwhile, it also incorporates the observation information of other traffic scheduling systems, so that the reinforcement learning model can learn the strategy of joint scheduling.
Compared with traditional traffic scheduling methods, it has the following advantages:
\begin{itemize}
\item We adopt reinforcement learning to learn the scheduling strategy, which is robust to the inaccuracy of demand prediction and adaptable in complex scheduling situations.
\item Compared with other scheduling schemes, it can take into account longer-term traffic demand, avoid local greed and achieve optimal scheduling over a long period of time.
\item JLRLS can realize joint scheduling among different traffic modes. When a certain traffic service is temporarily unavailable or inappropriate, more flexible traffic can be dispatched in time to meet the corresponding demand. The framework has strong scalability and can be applied to joint dispatch between buses on different routes. In the future, it
can also be applied to the more different joint scheduling.
\end{itemize}

\section{Overview}
In this section we define some concepts and notations used in the paper, and overview the framework of our model JLRLS.

\begin {table}[H]
\caption {Notations} \label{tab:title}
\begin{tabular}{c|l}
\hline
Notation & Description \\
\hline
$n$ 		&  The number of stations in the cluster \\
\hline
$m$ 		& The number of agents in system \\
\hline
$t$ 		& The $t$-th time segment in the future\\
\hline
$L$& The number of predicted time segments\\
\hline
$p$ & The longest time the passengers willing to wait\\
\hline
$s_i$ & The $i$-th station\\
\hline
$k$ & The feature dimension of the station in other systems\\
\hline
$w$ &  The feature dimension of the current system  \\

\hline
$O\in \mathbb{R}^{n\times k}$ & The observations for the stations in other systems\\
\hline
$H \in \mathbb{R}^w$ & The environmental factors of the current system \\
\hline
$G^t\in \mathbb{R}^{n\times n}$ & \tabincell{l}{The predicted flow network of bikes between stations
\\ in $t$-th time segment} \\
\hline
$l$ & The vector dimension after encoding $G$ \\
%
%
%
\hline
\end{tabular}
\end {table}
\subsection{Preliminary}

\begin{mydef}
\textbf{Agent}: Agent indicates buses in bus systems and the dispatching vehicles in bike-sharing systems.
\end{mydef}
\begin{mydef}
\textbf{Cluster}: Two types of cluster are defined for two different situations. For bus systems, the cluster is the bus stations sharing the same route.
For bike-sharing systems, the cluster represents the similar  stations, which are close to each other after clustering shown in section 3.1.2.
\end {mydef}

\begin{mydef}
\textbf{Demand}: We define two types of demand here. The first, demand for riding bikes, and taking buses from origin station to terminal. The second, demand for returning bikes, and taking buses from terminal to origin station.
\end{mydef}

\begin{mydef}
\textbf{Time segment}: Time segment is a period of time with fixed length, e.g, 15 mins.
\end{mydef}

\begin{mydef}
\textbf{Capacity}: Capacity stands for bus carrying capacity of passengers in bus systems, and vehicles for dispatch carrying capacity of bikes in bike-sharing systems.
\end{mydef}

\begin{mydef}
\textbf{Episode}: Episode is defined as a certain period in a day.
\end{mydef}

\subsection{Framework}
\begin{figure}[h]
  \centering
  \includegraphics[width=\linewidth]{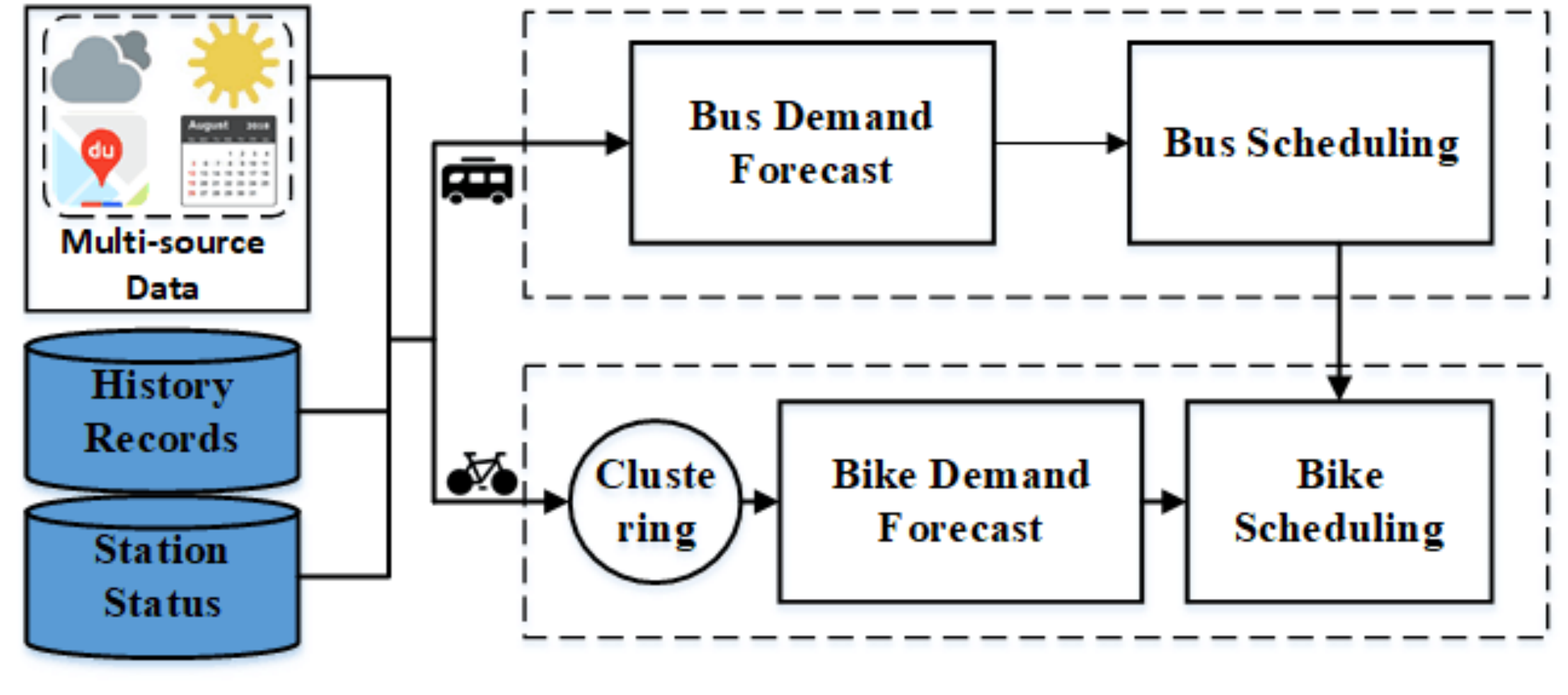}
  \caption{Framework of Joint Scheduling System}
 \label {fig:framework}
\end{figure}

We propose a general framework of Joint Long-term Reinforcement Learning Scheduling system (JLRLS). As shown in Fig \ref{fig:framework}, our model includes demand forecasting for stations and joint dispatching based on reinforcement learning. In bike-sharing scheduling, we incorporate  the information  of bus stations in bus scheduling system, so that the reinforcement learning model can learn the strategy of joint traffic scheduling.

\section{Method}


\begin{figure*}[h]
  \centering
  \includegraphics[height=5.5cm,width=\linewidth]{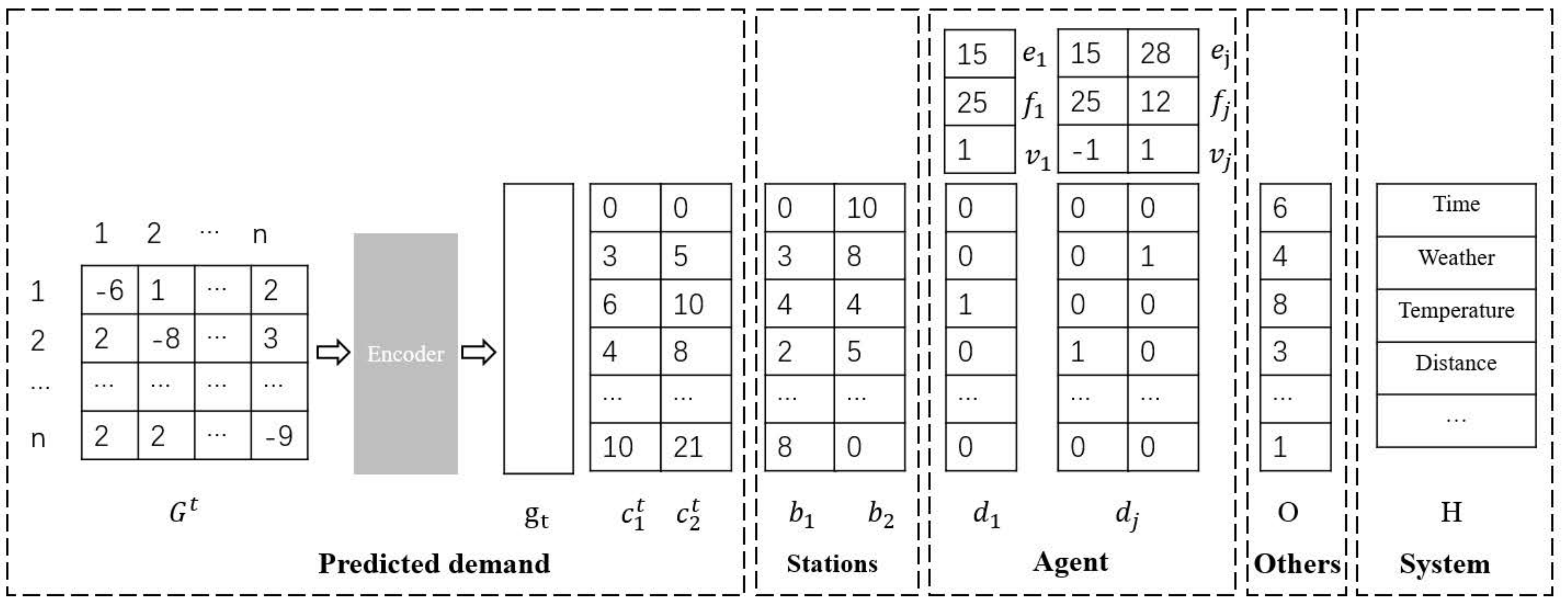}
  \caption{The State of System}
 \label {fig:state}
\end{figure*}
\subsection{Forecast System}

Since the characteristics are largely different between scenarios of bus systems and bike-sharing systems, we propose two kinds of prediction frameworks, bus flow forecast system and bike flow forecast system.
\subsubsection{Bus Flow Forecast System} \hfill

In bus flow forecast system, there is a relatively stable hierarchical concept, which is the passenger flow in each bus station is equal to that in the bus system. As passenger flow is regular and periodic in one day, daily total passenger flow of the bus system is also stable. We find that the scenario of bus passenger flow forecasting is similar to power system consumption forecasting. Inispired by it, we propose a bus flow prediction algorithm based on hierarchical time series.

Since the time series of daily passenger flow in a bus station exhibit strong regularity, in order to reduce the complexity of calculation,
we use linear model to learn the time series of the past time period for each station and the total bus system,
and predict the passenger flow in a short feature term. Because the individual forecast of the traffic at each station does not guarantee that their sum is consistent with the total flow of the bus system,
there is a summing matrix $S$ in the hierarchical time series forecasting to transform the whole problem into a regression problem which needs to be optimized.
\subsubsection{Bike Flow Forecast System} \hfill

Modeling the scene of bike-sharing flow is a very complex problem, because the time sequence in this scenario does not have strong regularity. In a bus system with fixed routes and stable users, many users only use shared bikes temporarily. Therefore, we might not use the method of bus flow forecast system. For the bike-sharing flow prediction system, our prediction model needs to predict the traffic flow between each station in the future. As a lot of stations sharing bikes, we implement \cite{li2018dynamic} to group the stations for simplifying the situation. The stations in each group are closer to each other and the traffic between them is more frequent than others. And we only consider the bike movement situation between the stations in each group.

Compared with traditional linear prediction algorithm, deep learning model like Long short-term memory~\cite{hochreiter1997long} (LSTM) is more suitable for modeling such unstable and nonlinear time series of bike-sharing station. Thus, we use LSTM to model the bike departure situation of each station in the short time of future.
 Specifically, $x$ indicate the time sequence of a bike station, the LSTM model maps an input sequence
${x_1, x_2, . . . , x_N }$ to outputs via a sequence of hidden states by computing the following equations recursively from $t = 1$ to $t = N$:
\begin{align*}
i_t &= \sigma(W_{xi}x_{t}+W_{hi}h_{t-1}+W_{ci}c_{t-1}+b_i),\\
f_t &= \sigma(W_{xf}x_{t}+W_{hf}h_{t-1}+W_{cf}c_{t-1}+b_f),\\
c_t &= f_{t}c_{t-1}+i_{t}tanh(W_{xc}x_{t}+W_{hc}h_{t-1}+b_c),\\
o_t &= \sigma(W_{xo}x_{t}+W_{ho}h_{t-1}+W_{co}c_{t}+b_o),\\
h_t &= o_{t}\textup{tanh}(c_t)
\end{align*}
where $x_t$, $h_t$ are the input and hidden vectors of the $t$-th time step,
$i_t$ , $f_t$ , $c_t$ , $o_t$ are the activation vectors of the input gate, forget gate,
memory cell and output gate, $W_{\alpha\beta}$ is the weight matrix between
vector $\alpha$ and $\beta$ (e.g, $W_{xi}$ is weight matrix from the input ${x_t}$ to the
input gate $i_t$ ), $b_\alpha$ is the bias term of $\alpha$ and $\sigma$ is the sigmoid function, and ${o_T}$ is the prediction of time series $x$.

Considering the movement of bikes between stations, when a bike leaves from station $A$, other stations in the same group may become the destination.
For this problem, we use the method of frequency replace probability to calculate the probability of bikes leave from station $A$ to other stations according to the past period.
Then the number of bikes predicted from station $A$ to station $B$ in the future period is the product of the total number of bikes predicted leave from station $A$ and the probability of station $A$ to $B$.

\subsection{Scheduling System}
After forecasting the bus flow and bike flow, we use reinforcement learning to produce the scheduling strategy, and adopt a Deep Deterministic Policy Gradient (DDPG) approach ~\cite{lillicrap2015continuous}.
It incorporates the information of bike flow to avoid local greed. At the same time, it also incorporates the observation information of other traffic scheduling systems, so that the reinforcement learning model can learn the strategy of traffic joint scheduling.

\subsubsection{The State of Scheduling System}\hfill

In order to describe the state clearly, we summarize it in Figure~\ref{fig:state}.
As we can see, the state is divided into five categories: predicted demand, station information, agent states, scheduling information of other traffic systems and the state of system.
There are a matrix $G^t \in \mathbb{R}^{n \times n}$ and three vectors $g^t \in \mathbb{R}^l, c_1^t, c_2^t \in \mathbb{R}^n$ in the predicted demand, where $G_{ij}^t$ represents the number of bikes from $s_i$ to $s_j$ in the $t$-th time segment. $g^t$ represents vector after encoding $G^t$.
$c_1^t, c_2^t \in \mathbb{R}^n$ respectively indicate the first and the second type of demand for each station in the $t$-th time segment.
Both $b_1$ and $b_2 \in \mathbb{R}^n$ denote station information. In bus systems, they represent the time interval of the most recent bus going forward and backward at a station, respectively.
In bike-sharing systems, $b_1$ means the available bikes, and $b_2$ is the available docks at the station.
$d_1 \in \mathbb{R}^n$ is an one-hot vector, which stands for which station the current agent will locate at.
$e_1 ,f_1\in \mathbb{R}$ respectively stand for the capacity occupied and the remaining capacity for current agent.
$v_1 \in \mathbb{R}$ represents the operation being performed.
In bus systems, there are three types of operations, $v_1 \in$ \{-1,0,1\}, -1 standing for driving from $s_1$ to $s_n$, 0 for halting, and 1 standing for driving from $s_n$ to $s_1$.
In bike-sharing systems, $v_1$ means how many bikes are loaded or unloaded, $v_1$ > 0 for loading, $v_1$ < 0 for unloading, and $v_1$ = 0 standing for not moving.
$d_j,e_j,f_j,v_j$ respectively stand for the corresponding state of other agents.
 $O$ can be the observation of the bus system by bike-sharing system, or it can be the observation of the bus system on different routes.
$H $ represents the environmental factors such as the weather, temperature, distance.
\subsubsection{Bus Scheduling System} \hfill

\textbf{A State.} For bus scheduling system, we define the state as follows:
\begin{itemize}
\item Observation for bus stations, $(b_1, b_2, c_1^1, c_2^1, ..., c_1^L, c_1^L).$
\item The state of the scheduled bus, $(d_1, e_1, f_1, v_1).$
\item The state of other buses on the same route, $(d_j, e_j, f_j, v_j).$
\item Observation for the system, ($H$).
\item Observation for stations in other traffic systems, ($O$).
\end{itemize}


\textbf{An Action.} For a bus, there are three types of an action:

 (1) towards to terminal $s_n$; (2) toward to origin station $s_1$; (3) stoping at $s_1$ or $s_n$.

\textbf{A Reward.} We set the reward mechanism as follows: (1) Each time the bus travels from a to b, the reward is the reduced waiting time, where the punishment is related to the driving time; (2) The bus stops driving, no rewards and punishments.

\textbf{Stopping Condition.}
A passenger waiting for p time segments or an episode is completed.

\subsubsection{Bike Scheduling System}\hfill

\textbf{A State.}
For a bike-sharing scheduling system, the state consists of the following five parts:
\begin {itemize}
\item Observation for bike-sharing stations, ($b_1, b_2, c_1^1, c_2^1, \cdots,  c_1^L,\\ c_2^L, g^1,\cdots,g^L$).
\item State of the current dispatch vehicle, $(d_1, e_1, f_1, v_1).$
\item State of other vehicles for dispatch in the same cluster, $(d_j, e_j, f_j,\\ v_j)$.
\item Observation for the system, ($H$).\
\item Observation for bike-sharing stations in other traffic systems ($O$), such as $(b_1, b_2)$ in the bus system.
\end {itemize}

Different from  \cite{li2018dynamic}, we consider more detailed information on the flow of bikes between stations.
We use $G^t$ to represent the predicted flow network of bikes between stations in the future. The matrix $G^t$ will result in a high complexity,
so we encode the matrix $G^t$ to get a vector $g^t $, which keeps the bike flow  between stations. It comprehensively describes the bike flow network,
and simplifies the representation of the state, which is more conducive to the convergence of the strategy and the exploration of the agent. To enable the bike-system working with the bus system, we incorporate the observation information of bus scheduling systems in reinforcement learning.

\textbf{An Action.}
An action is defined as $(d_1, v_1)$. $d_1$ denotes which station the current dispatch vehicle will unload or load bikes, and $v_1$ denotes the number of unloaded or loaded bikes.

\textbf{Reward.}
After an episode is completed, we set the reward mechanism as follows: $(1)$ we reward the agent as the number of services provided by bike; $(2)$ The punishment is related to the cost of scheduling and the number of bikes exceeding total capacity.

\textbf{Stop Condition.} When an episode is completed.
    
Our model has the following advantages over \cite{li2018dynamic}:
\begin{itemize}
\item The representation of the state contains more detailed flow information between stations in the future, which is more conducive to policy convergence.
\item We adopt a Deep Deterministic Policy Gradient (DDPG) approach. First of all it is an Actor-Critic network, taking into account the advantages of Value-Based and Policy-Based methods. Secondly, using LSTM inside the Actor network, it can comprehensively consider the historical information of the state.
\item We consider the interactions between different traffic systems, so as to jointly dispatch different traffic systems and improve traffic efficiency.
\end{itemize}

\section{CONCLUSIONS}
In order to provide a better travel experience, we urgently need a joint scheduling system capable of jointly schedule multiple modes of transportation. Therefore, we propose the above topic and give our solution. To successfully complete this research, we need more resource, including but not limited to the following:

\begin{enumerate}
\item Complete query records of Baidu map App.
\item The bicycle histories of Baidu partners.
\item The routes of buses and the passenger flow at different time.
\item Enough GPU resources.
\end{enumerate}
Multi-modal scheduling is an indispensable part of smart city. The
successful development of multi-modal transportation scheduling
could make a lots of advantages, such as reducing transport times, balancing traffic flows, reducing traffic congestion, and ultimately, improving efficiency of intelligent transportation systems.
Therefore, the research of our topic is valuable to the project of smart city. We believe that after possessing these resources we can develop a more comprehensive and efficient multimodal joint scheduling system.


\bibliographystyle{unsrt}
\bibliography{samplebase}

\appendix

\end{document}